\pdfoutput=1

\documentclass[12pt]{article}

\usepackage{acl}
\usepackage{tabularx}
\usepackage{float}

\usepackage{times}
\usepackage{latexsym}

\usepackage{graphicx}

\usepackage[T1]{fontenc}
\usepackage{array}

\usepackage[utf8]{inputenc}
\usepackage{hyperref}
\usepackage{microtype}

\title{Enhancing Multilingual Language Models for Code-Switched Input Data}

\author{
  Katherine Xie  \\
  MIT \\
  \texttt{kxie22@mit.edu} \\
  \And
  Nitya Babbar \\
  MIT  \\
  \texttt{nityab@mit.edu} \\
  \And
  Vicky Chen \\
  MIT \\
  \texttt{vickywc@mit.edu} \\
  \And
  Yoanna Turura \\
  MIT \\
  \texttt{yturura@mit.edu} \\
}

\begin{document}

\maketitle

\section{Abstract}
Code-switching, or alternating between languages within a single conversation, presents challenges for multilingual language models on NLP tasks. This research investigates if pre-training Multilingual BERT (mBERT) on code-switched datasets improves the model's performance on critical NLP tasks such as part of speech tagging, sentiment analysis, named entity recognition, and language identification. We use a dataset of Spanglish tweets for pre-training and evaluate the pre-trained model against a baseline model.

Our findings show that our pre-trained mBERT model outperforms or matches the baseline model in the given tasks, with the most significant improvements seen for parts of speech tagging. Additionally, our latent analysis uncovers more homogenous English and Spanish embeddings for language identification tasks, providing insights for future modeling work.

This research highlights potential for adapting multilingual LMs for code-switched input data in order for advanced utility in globalized and multilingual contexts. Future work includes extending experiments to other language pairs, incorporating multiform data, and exploring methods for better understanding context-dependent code-switches.

\section{Introduction}

In today’s increasingly globalized world, multilingualism is common, and code-switching—switching between languages within a single conversation—is particularly prevalent in informal communication. However, most current natural language processing (NLP) models are designed for monolingual text or cross-lingual tasks such as translation, leaving a significant gap in handling code-switched text. Code-switching presents unique challenges, including accurately capturing language transitions and seamlessly determining meaning across them. As demand grows for multilingual and mixed-language processing in applications such as chatbots and social media monitoring, addressing these challenges is critical, especially for multilingual and multicultural communities.

Existing multilingual language models (LMs), such as mBERT (Multilingual BERT) and XLMR (Cross-lingual Language Model), are typically pre-trained on monolingual or translation-based datasets \cite{conneau2020unsupervised}, making them ill-suited for handling the unpredictable language transitions found in code-switched text. This limitation affects various NLP related tasks, often resulting in higher loss values and lower accuracies than monolingual inputs.

Our study aims to fill this gap by pre-training mBERT, an existing multilingual LM, on a Spanglish code-switched dataset and focusing on evaluating the model with four different tasks: parts of speech tagging (POS), language identification (LID), named entity recognition (NER), and sentiment analysis (SA) tasks. In addition, we investigate the differences in latent space representations generated by the model before and after pre-training to interpret how the model's understanding of code-switched input data changes. By improving the model's ability to capture language transitions and generate high-quality outputs from code-switched text, we aim to contribute not only to technical advances in NLP but also to provide better natural language understanding in multilingual communities.

\section{Related Work}
Research in code-switching within the field of natural language processing has gained traction in recent years due to the challenges of handling multilingual data, including capturing linguistic transitions, syntactic patterns, and semantic coherence across different languages. 

Although not specifically designed for code-switching, a paper by \citet{ammar2016massively} laid some groundwork for multilingual models by introducing embeddings that allow words from different languages to be represented in a shared vector space. This has led to the development of multilingual models like mBERT and XLMR, which in recent years have been applied towards cross-lingual tasks. A study by \citet{aguilar2020lince} developed the LinCE benchmark to provide standardized evaluation metrics for code-switching across various tasks and language pairs, aiming for consistency in evaluating the performance of multilingual models on code-switched data. 

A study by \citet{roberta2019} introduced RoBERTa, an optimized variant of BERT, which refined BERT’s training process to improve performance without task-specific fine-tuning. RoBERTa removed the Next Sentence Prediction (NSP) objective included in the original BERT model, but was found to add limited value to downstream tasks. The researchers also implemented dynamic masking, where the tokens selected for masking were randomized in each training iteration, enhancing the model’s ability to generalize instead of overfitting to the fixed masked word positions. Although modifications were made on BERT, these changes have potential to improve the performance of mBERT on code-switched data to better capture complex linguistic patterns across languages.

In 2018, \citet{bhat2018universal} contributed to research on code-switched language modeling by proposing models specifically tailored for Hindi-English code-switched data, using LSTM-based approaches, which proved effective for sequence labeling tasks but were limited in other abilities. 

There has also been research related to latent space representations created by models from code-switched data. A study by \citet{wu} uses Latent Language Space Models (LLSMs) to focus on accurately identifying language change points within code-switched data, while \citet{asnani} use Latent Dirichlet Allocation (LDA), which is an extension of Probabilistic Latent Semantic Analysis (PLSA), to focus on topic modeling within code-switched text. A paper by \citet{harish} includes interpretability analysis on latent space embeddings generated by pre-trained models on code-switched text.

Outside of the field of natural language processing, many research papers have focused on the linguistics of code-switching. Work by \citet{lipski2008} and \citet{dussias} focused on exploring grammatical rules for code-switching, and where language transitions most frequently occur within a sentence. These results can be used to test a model's understanding of grammatical rules related to code-switching.

Our work builds upon this existing research by directly addressing the limitations of current multilingual models in handling NLP tasks in code-switched contexts. We leverage the mBERT model, fine-tune it on code-switched datasets, and conduct experiments to analyze how the new model understands code-switched data. We aim to improve upon the model’s ability to produce accurate output based on multilingual sentences, and gain some intuition as to how the improved model captures their semantic meaning and whether this aligns with generally accepted grammatical rules for code-switching.

\section{Evaluation}
We will evaluate our pre-trained model for its performance on token classification and sentence classification tasks, as measured by various metrics. The model’s performance will be benchmarked against a baseline mBERT transformer model without specific code-switching pre-training.

We will use the LinCE benchmark \cite{aguilar2020lince}, which provides annotated datasets for tasks such as POS tagging, LID, NER, and SA tasks across several language pairs. We will use the provided Spanglish datasets for each of these tasks that contain short code-switched tweets. Performance of our model will be evaluated using LinCE’s accuracy, F1 score, Precision, and Recall metrics for each task. 

Additionally, for the SA and LID tasks, we use latent semantic analysis to observe how the latent space embeddings of code-switched input data differ before and after pre-training. We compress the representations to two dimensions to allow for visual analysis through graphing.

Finally, for the POS tagging task, we analyze the parts of speech on which code-switching most frequently occurs within a sentence. We then compare this data to linguistic literature such as \citet{lipski2008} and \citet{dussias} to determine whether our model seems to accurately capture the grammar rules of code-switching.

In the sections below, we outline each of the tasks we used to evaluate our model. Evaluating LMs on these four tasks with code-switched text can help to determine the model’s ability to process multilingual data and is a step toward more advanced tasks like text generation and machine translation in code-switched environments.

\subsection{Language Identification}

One of our tasks focuses on language identification (LID) in Spanish-English code-switched data. This dataset includes bilingual tweets, where each word has a corresponding label.

Example of a single document from the LID dataset:

["I", "really", "quiero", "comer", "tacos", "tonight", "!"]

May have the following label:

["B-LANG1", "B-LANG1", "B-LANG2", "B-LANG2", "B-LANG2", "B-LANG1", "O"]

Non-linguistic words or characters outside of either language classification are labeled with “O”, for other.

\subsection{Part of Speech Tagging}

Part-of-speech (POS) tagging is another NLP task that involves labeling each word in a sentence with its part-of-speech, such as noun, verb, or adjective. This task is crucial for language models to capture syntactic and semantic relationships between words. Code-switching contexts present the POS tagging task with more challenges because there can be a change in the grammatical roles or structure of a sentence mid-way through, requiring the model to accurately adapt to the change. 

Example of a single document from the POS dataset: 

["er", "los", "Burger", "Kings", "and", "McDonalds", "todo", "eso", "para", "allá", "es", "."]

May have the following label denoting the part of speech of every token:

["INTJ", "DET", "PROPN", "PROPN", "CONJ", "PROPN", "ADJ", "PRON", "ADP", "ADV", "VERB", "PUNCT"].

\subsection{Sentiment Analysis}
Sentiment analysis (SA) is a natural language processing technique that categorizes text by sentiment. In this case, the categories are positive, negative, or neutral.

Example of a single document from the SA dataset:

["Maury", "bien", "happy", "with", "his", "present", "que", "le", "regalo", "su", "Padrino", "Oscar", "!", "\#futuremusico"]

Has a \textit{positive} label because the overall sentiment of this sentence is positive.

\subsection{Named Entity Recognition}
Named entity recognition (NER) is the fourth NLP task we chose to experiment with as it involves identifying and  classifying specific words or phrases within a text as belonging to predefined categories (entities). In code-switching contexts, this task is a unique challenge as the model has to learn to maintain context and relevance across language boundaries and improve multilingual understanding and reduce misclassification.

Example of a single document from the NER dataset: 

["Im", "so", "ready", "for", "Camelia", "la", "Tejana"]

Has the following labels:

["O", "O", "O", "O", "B-TITLE", "I-TITLE", "I-TITLE"]

\section{Methodology}
To implement a language model for code-switching text, we will leverage a variety of open-source tools and datasets designed for multilingual processing. Our primary evaluation benchmark will be on part-of-speech tagging, language identification, named entity recognition, and sentiment analysis tasks.

The foundation of our language model will be from the Hugging Face Transformers library, which supports a wide range of pre-trained models for language tasks, including the multilingual model mBERT. mBERT has been pre-trained on a large multilingual corpora of 150+ languages. To implement our model, we will be fine-tuning mBERT on Spanglish code-switched data with a masked language modeling (MLM) task to capture the linguistic intricacies present in mixed-language sentences.

The training process will involve using cloud-based resources like Google Colab GPUs to efficiently fine-tune and evaluate the models on code-switched datasets. Our datasets will also be kept in cloud-based storage.

Before making any modifications, we first evaluate the performance of the original mBERT model on each of our chosen NLP tasks in order to obtain a starting benchmark. We used the LinCE corpus by \citet{aguilar2020lince}, which contains Spanish-English code-switched datasets. In the sections below, we outline the specific approaches to each step in our evaluation pipeline, along with any challenges we encountered.

\begin{figure}
    \centering
    \includegraphics[width=0.9\linewidth]{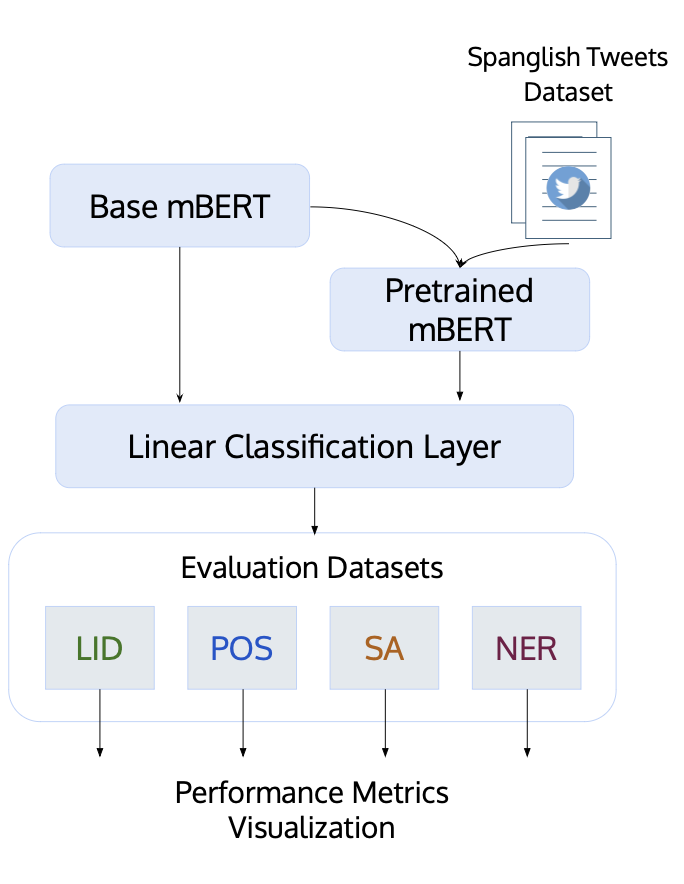}
    \caption{Pipeline for evaluating the impact of fine-tuning mBERT on code-switched data}
    \label{fig:pipeline}
\end{figure}

The general pipeline of our study is shown in \autoref{fig:pipeline}.

\subsection{Pretraining mBERT}
We fine-tuned the mBERT model on a Spanglish dataset using an unsupervised task of masked language modeling (MLM). The dataset, comprised of Spanglish sentences, was preprocessed by removing leading and trailing whitespace, filtering out empty lines, and converting it into a Hugging Face Dataset object. The text was then tokenized using the mBERT \texttt{bert-base-multilingual-cased} tokenizer with padding and truncation with the max token length set to 128. Special tokens such as [MASK] were preserved to support the MLM objective. We employed a dynamic masking strategy similar to that of RoBERTa \cite{roberta2019} where 15\% of tokens were randomly masked in each batch using the DataCollatorForLanguageModeling from Hugging Face. 

For pretraining, the MLM head of mBERT was fine-tuned on the tokenized dataset using the Hugging Face Trainer API. The model was trained for 3 epochs with a batch size of 8 per device with. The training was performed on a single GPU, and the fine-tuned model and tokenizer were saved for later use. The pipeline was implemented using Python and the Hugging Face Transformers library.

\subsection{Token Classification Tasks}
The first step in the evaluation pipeline for token classification tasks including POS, LID, and NER involved preprocessing datasets formatted in the CoNLL standard. Sentences and corresponding tags for each token were parsed and stored into lists. The unique tags were mapped to integer IDs. Input sentences and tags were tokenized using the Hugging Face tokenizer with a maximum sequence length of 64 with padding and truncation if needed. Non-aligned tokens were assigned -100 to ensure they were ignored during training. The data was then converted into PyTorch datasets using a custom TokenClassificationDataset class that stores input IDs, attention masks, token type IDs, and labels as tensors. 

Either the pretrained or baseline mBERT model was then loaded using the Hugging Face AutoModelForTokenClassification. A linear classification layer was automatically added, and the base model layers were frozen to focus on training this classifier head. For training, the Hugging Face Trainer API was used with hyperparameters: a learning rate of $2 \times 10^{-5}$, a batch size of 16, 3 training epochs, and a weight decay of 0.01. Metrics such as accuracy, precision, recall, and F1 score were computed using the seqeval library. 

\subsection{Sentence Classification Tasks}
The evaluation pipeline for sentence classification tasks (e.g. sentiment analysis) began by preprocessing a dataset formatted in the CoNLL-style. Sentences and corresponding sentence labels are parsed. Sentence labels were mapped to integer IDs. The dataset was then converted to a custom dataset class that tokenized the input sentences with a Hugging Face tokenizer. The tokenization applied a maximum sequence length of 128, with padding and truncation if needed. The data was then converted to PyTorch tensors, including input IDs, attention masks, and integer labels. 

For model training, a custom classifier class (mBERTWithClassifier) was built from either our pretrained mBERT or baseline mBERT model, using the pooled output from the [CLS] token as the input to a linear classification head. The classifier head mapped the hidden state dimensions to the number of unique sentence labels. Loss was computed using a cross-entropy function. Models were trained using the Hugging Face Trainer API with hyperparameters: a learning rate of $2 \times 10^{-5}$, a batch size of 16, 3 training epochs, and a weight decay of 0.01. Performance metrics were computed using the sklearn library. 

\subsection{Latent Space Analysis}
To evaluate linguistic constraints in the learned representations, we conducted our latent space analysis (LSA) using the base multilingual BERT model and our fine-tuned version of the same base mBERT on the code-switched Spanglish dataset. 

Our latent space representations were from English and Spanish words found in the LinCE Language Identification dataset. This code-switched dataset was annotated with word-level language labels, specifically distinguishing between English (en) and Spanish (es). These annotations allowed us to group words by language to facilitate downstream analysis. 

After splitting the data into two categories, we tokenized English and Spanish words using the shared tokenizer from the Hugging Face transformers library. For each word, we extracted its embedding by passing it through the model and isolating the vector corresponding to the [CLS] token. This representation was chosen as it captures the sentence-level semantics, making it suitable for analyzing latent space clustering. For each model, embeddings were computed for 50 randomly sampled English words and 50 randomly sampled Spanish words. 

To visualize the high-dimensional embeddings, we employed Uniform Manifold Approximation and Projection (UMAP), a nonlinear dimensionality reduction technique well-suited for maintaining both local and global structures in the data. The embeddings were projected onto a two-dimensional space for easier interpretability. 
UMAP parameters were set as follows: 
\begin{itemize}
    \item Number of Neighbors: 5
    \item Minimum Distance: 0.3
    \item Metric: Cosine similarity
\end{itemize}

\section{Results and Analysis}

Evaluation metrics include accuracy, F1 score, precision, and recall, which allow us to assess the base model’s performance on both general accuracy and the ability to handle ambiguous tokens. Our results are shown in the table in \autoref{fig:table}. This highlights how it performs well on tasks like POS tagging as it is able to leverage predictable grammatical patterns in code-switched texts. It shows moderate performance in language identification when handling ambiguous tokens. However, it struggles with more context-dependent tasks like NER where the F1 score is near 0. This can also be attributed to the fact that the labels in the dataset used for evaluating NER tasks are largely `O' labels, so this task may benefit from re-weighting techniques that help the model better recognize other NER tags.

\begin{figure}
\small
    \centering
    \begin{center}
    \begin{tabular}{ |p{1.0cm}|c|c|c|c| }
    \hline
    \textbf{Metric} & \textbf{LID} & \textbf{POS} & \textbf{SA} & \textbf{NER} \\ 
    \hline
    Loss & 0.964500 & 0.681921 & 0.9769 & 0.1336 \\ 
    \hline
    Accuracy & 80.73\% & 85.71\% & 56.48\% & 97.59\%
    \\
    \hline
    F1 Score & 0.2825 & 0.8067 & 0.5442 & 0.0031\\
    \hline
    Precision & 0.2541 & 0.7997 & 0.5361 & 0.5556\\
    \hline
    Recall & 0.3179 & 0.8137 & 0.5648 & 0.0015\\
    \hline
    \end{tabular}
    \caption{Validation Results on NLP Tasks Using LinCE Benchmark}
    \label{fig:table}
    \end{center}
\end{figure}

\begin{figure}
    \centering
    \includegraphics[width=1.00\linewidth]{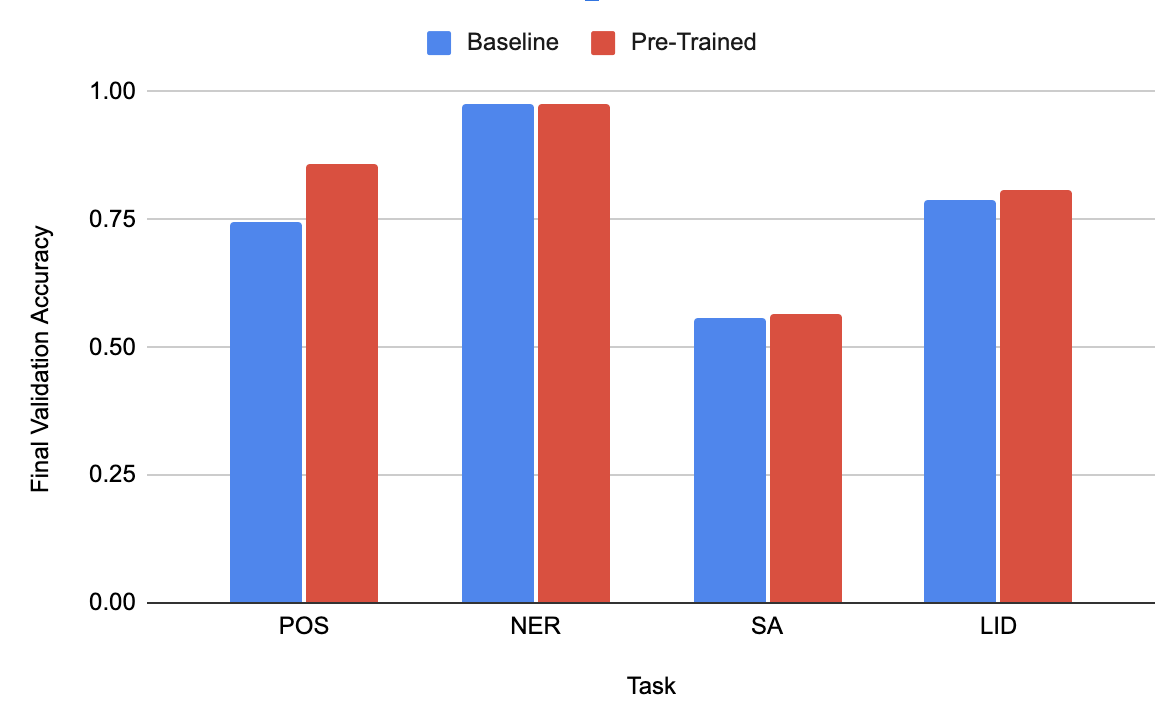}
    \caption{Pre-trained vs. Baseline  mBERT Model Validation Accuracy}
    \label{fig:accgraph}
\end{figure}

The bar graph in \autoref{fig:accgraph} compares the performance between the baseline mBERT model and our pre-trained model for each of the 4 tasks (POS, NER, SA, and LID). The pre-trained model outperforms or at least matches the accuracy of the baseline for all tasks, demonstrating improvements from fine-tutuning on a code-switched dataset. POS tagging shows the greatest improvement, suggesting that structural linguistic features could benefit from this pre-training. SA and LID also show improvements, although less pronounced than in POS. This is likely due to the more context-dependent nature of those tasks. For NER, the minimal performance gap suggests the task may be inherently well-suited to the baseline model.

\begin{figure}
    \centering
    \includegraphics[width=1.00\linewidth]{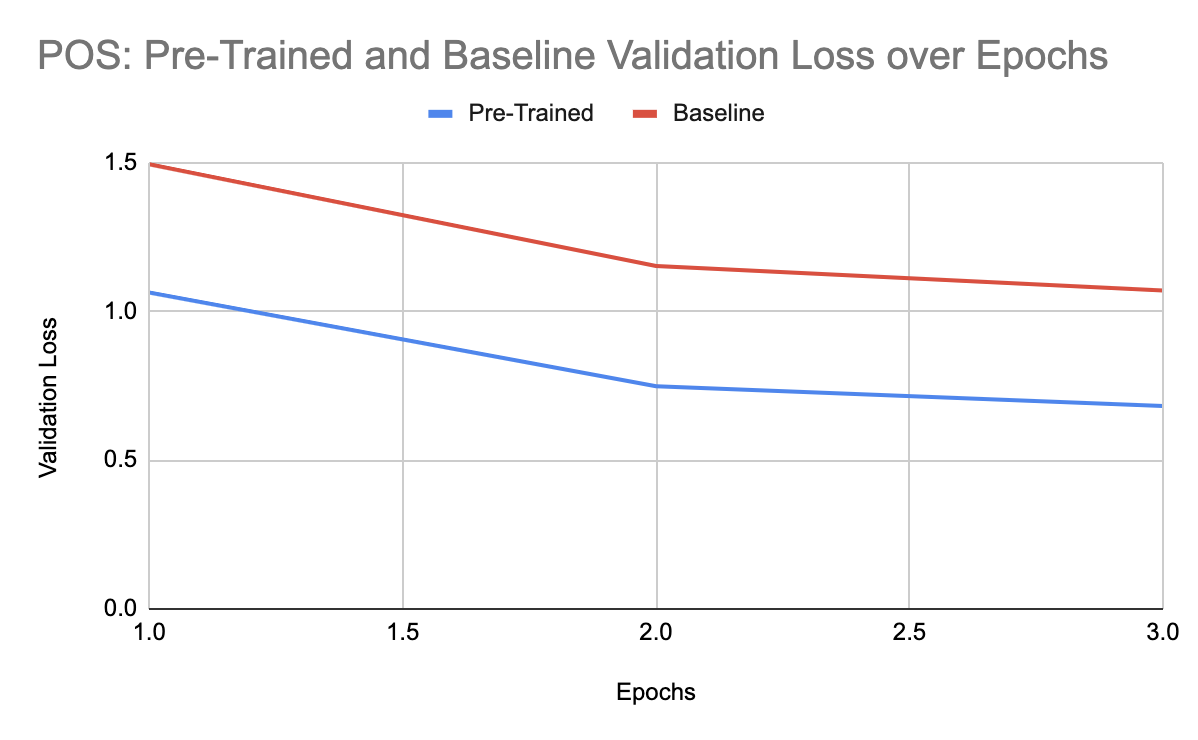}
    \caption{Pre-trained vs. Baseline  mBERT Validation Loss on POS Task}
    \label{fig:lossgraph}
\end{figure}

\autoref{fig:lossgraph} compares the validation loss over training epochs for POS tagging between the baseline and pre-trained mBERT models. Our pretrained has consistently lower loss over the 3 epochs. This indicates its ability to better generalize to unseen code-switched data. Additionally, the mostly steady decline in loss for our pre-trained model suggests that through the fine-tuning, the model is adapting more effectively to the nuances of code-switching, whereas the baseline model appears to show slower improvements and an overall higher loss throughout the training process.

\begin{figure}
    \centering
    \includegraphics[width=1.00\linewidth]{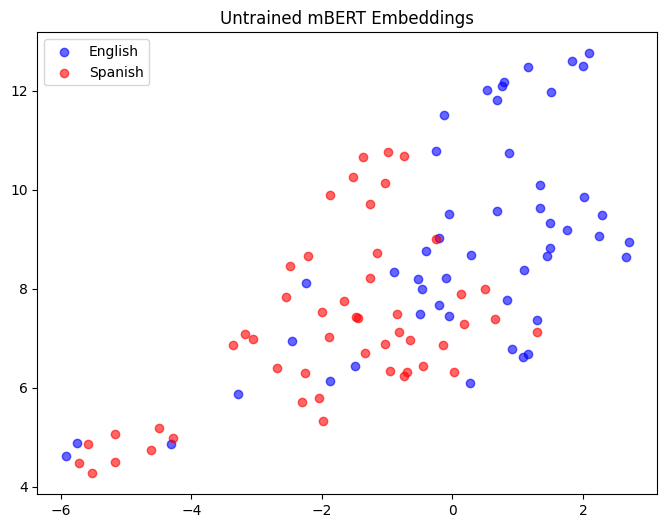}
    \includegraphics[width=1.00\linewidth]{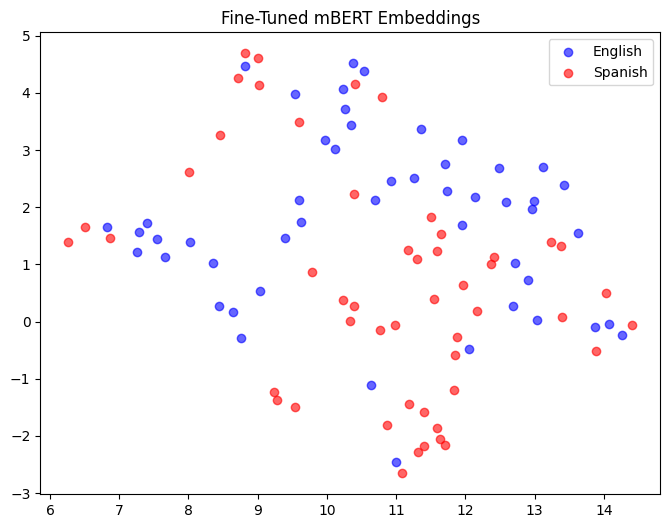}
    \caption{UMAP visualization of English and Spanish word embeddings before and after fine-tuning on code-switched data.}
    \label{fig:lsa1}
\end{figure}

We also visually analyzed our UMAP plots of the latent embeddings (\autoref{fig:lsa1}) from these models. Before fine-tuning, English and Spanish word embeddings were more distinctly separated, reflecting language-specific clusters in the untrained mBERT. After fine-tuning on code-switched data, the embeddings became more scattered within the same region, indicating a higher degree of homogeneity between English and Spanish embeddings. This suggests that the pre-trained model may be better capturing the underlying semantic meaning of input tokens, rather than the more ``superficial" aspect of language in its hidden representations.

\section{Conclusion}

Code-switching is increasingly common in multilingual communities, but current language models struggle to handle mixed-language text effectively. Our research has aimed to improve multilingual BERT by fine-tuning on the code-switched Spanglish dataset. We demonstrate the potential of fine-tuning multilingual models on code-switched data, revealing significant benefits for tasks like POS tagging while also highlighting the challenges of tasks like Sentiment Analysis in this context. The results underscore the importance of task-specific considerations when leveraging multilingual models for code-switching applications, as well as the inherent strengths and limitations of baseline pre-trained models like mBERT.

Beyond task performance, our latent space analysis provided deeper insights into how fine-tuning impacts embeddings. We found that pre-training allows the model to better capture the fluid transitions characteristic of code-switching, creating a shared representational space that aligns with the linguistic dynamics of multilingual, code-switched data.

Future work can extend these findings by scaling experiments to other code-switching datasets, such as Mandarin-English or Arabic-French, to assess the scalability and generalizability of these observations. Incorporating multimodal data, such as images or audio paired with text, could provide further insights into how additional cues influence code-switching behavior. Additionally, exploring context-dependent phenomena, such as changes in topic, formality, or emotional tone, could help capture the underlying drivers of code-switching. Finally, applying alternative methods to evaluate latent space representations could offer more nuanced interpretations of how multilingual models adapt to the complexities of code-switched data.

\section{Impact Statement}
Our project, fine-tuning models on code-switched data to improve their understanding of multilingual text, has the potential for increasing inclusivity and representation in the field of natural language processing (NLP). Code-switched text presents unique challenges, so this work provides a pipeline for creating language models that can better serve multilingual and code-switching populations. While not much extensive research has been done on language models for code-switching data, there are many real-world applications of code-switching language models including chatbots, virtual assistants, social media monitoring, and more. Moreover, the model's ability to handle linguistic transitions between languages can help to overall bridge communication divides in informal or culturally mixed situations. As such, our project will contribute to the development of NLP models and technologies that focus on equity and access for underrepresented communities where code-switching may be prevelant.

Although there are many benefits and potential applications of this research, there are also many important ethical and societal considerations to be made. Many NLP tasks that models are trained on use predefined labels in their datasets. Since these labels are often discrete, they carry the risk of oversimplifying linguistic nuances, particularly in spectrum-like tasks. For instance, sentiment analysis often simply uses labels such as ``positive", ``negative", and ``neutral"; however, sentiment is not usually discrete, so labels such as ``mildly positive" or ``very negative" may better capture the spectrum-like nature of sentiment analysis. Another example is detecting code-switching, where certain words shared by both languages may not equally belong to each—depending on the context, a word might be 70\% English and 30\% Spanish rather than an even 50-50 split. These predefined labels can lead to overgeneralization across categories and languages that can fail to accurately capture the complexities of code-switched text. 

Furthermore, implementing these models in real-world scenarios raises privacy and equitable resource distribution concerns. Social media monitoring, for instance, could involve training models on user-generated content without explicit consent, risking violations of privacy. While this work addresses widely used code-switched languages like Spanglish, it could inadvertently divert resources away from lesser-represented languages or dialects that also lack digital representation, further marginalizing certain communities. This consequence can risk exacerbating power imbalances, granting more technological advantages to already dominant demographic groups. To mitigate these risks, it is important to implement ethical data collection practices, conduct dataset audits to minimize biases, and ensure continuous evaluation for fairness and accountability in model deployment. By emphasizing transparency, fairness, and inclusivity in code-switched model development, we want to set a precedent for ethical AI model development while advancing the studies of multilingual NLP.


\begin{thebibliography}{10}
\expandafter\ifx\csname natexlab\endcsname\relax\def\natexlab#1{#1}\fi

\bibitem[{Aguilar et~al.(2020)Aguilar, Kar, and Solorio}]{aguilar2020lince}
Gustavo Aguilar, Sudipta Kar, and Thamar Solorio. 2020.
\newblock Lince: A centralized benchmark for linguistic code-switching evaluation.
\newblock In \emph{Proceedings of The 12th Language Resources and Evaluation Conference}, pages 1803--1813.

\bibitem[{Ammar et~al.(2016)Ammar, Mulcaire, Tsvetkov, Lample, Dyer, and Smith}]{ammar2016massively}
Waleed Ammar, Phoebe Mulcaire, Yulia Tsvetkov, Guillaume Lample, Chris Dyer, and Noah~A Smith. 2016.
\newblock Massively multilingual word embeddings.
\newblock In \emph{arXiv preprint arXiv:1602.01925}.

\bibitem[{Asnani and Pawar(2018)}]{asnani}
Kavita Asnani and Jyoti Pawar. 2018.
\newblock Extraction of code-mixed aspect topics in semantic representation.

\bibitem[{Bhat et~al.(2018)Bhat, Bhat, Sharma, and Bhatt}]{bhat2018universal}
Irshad Bhat, Riyaz~Ahmad Bhat, Dipti~Misra Sharma, and Rajesh Bhatt. 2018.
\newblock Universal dependency parsing for hindi-english code-switching.
\newblock In \emph{Proceedings of the 2018 Conference of the North American Chapter of the Association for Computational Linguistics: Human Language Technologies}, pages 987--998.

\bibitem[{Conneau et~al.(2020)Conneau, Khandelwal, Goyal, Chaudhary, Wenzek, and Guzman}]{conneau2020unsupervised}
Alexis Conneau, Kartikay Khandelwal, Naman Goyal, Vishrav Chaudhary, Guillaume Wenzek, and Francisco Guzman. 2020.
\newblock Unsupervised cross-lingual representation learning at scale.
\newblock \emph{arXiv preprint arXiv:1911.02116}.

\bibitem[{De~Leon et~al.(2024)De~Leon, Madabushi, and Lee}]{harish}
Frances De~Leon, Harish Madabushi, and Mark Lee. 2024.
\newblock Code-mixed probes show how pre-trained models generalise on code-switched text.

\bibitem[{Dussias et~al.(2014)Dussias, Tamargo, Kroff, and Gerfen}]{dussias}
Paola Dussias, Rosa Tamargo, Jorge Kroff, and Chip Gerfen. 2014.
\newblock Looking into the comprehension of spanish-english code-switched sentences: Evidence from eye movements.

\bibitem[{Lipski(2008)}]{lipski2008}
John Lipski. 2008.
\newblock Spanish, english, or... spanglish?

\bibitem[{Liu et~al.(2019)Liu, Ott, Goyal, Du, Joshi, Chen, Levy, Lewis, Zettlemoyer, and Stoyanov}]{roberta2019}
Yinhan Liu, Myle Ott, Naman Goyal, Jingfei Du, Mandar Joshi, Danqi Chen, Omer Levy, Mike Lewis, Luke Zettlemoyer, and Veselin Stoyanov. 2019.
\newblock Roberta: A robustly optimized bert pretraining approach.

\bibitem[{Wu et~al.(2015)Wu, Shen, and Hsu}]{wu}
Chung-Hsien Wu, Han-Ping Shen, and Chun-Shan Hsu. 2015.
\newblock Code-switching event detection by using a latent language space model and the delta-bayesian information criterion.

\end{thebibliography}
\end{document}